\documentclass{article}
\pdfoutput=1
\PassOptionsToPackage{numbers, compress}{natbib}
\usepackage[preprint]{neurips_data_2022}
\usepackage{booktabs}
\usepackage[table,xcdraw]{xcolor}
\usepackage{hyperref}
\usepackage[utf8]{inputenc} % allow utf-8 input
\usepackage[T1]{fontenc}    % use 8-bit T1 fonts
\usepackage{hyperref}       % hyperlinks
\usepackage{url}            % simple URL typesetting
\usepackage{booktabs}       % professional-quality tables
\usepackage{amsfonts}       % blackboard math symbols
\usepackage{nicefrac}       % compact symbols for 1/2, etc.
\usepackage{microtype}      % microtypography
\usepackage{xcolor}         % colors
\usepackage{graphicx}
\usepackage{subfig}
\usepackage{todonotes}
\usepackage{authblk}
\usepackage{tabularx}
\usepackage{enumitem}
\makeatletter
\def\input@path{{./sections/}}
\makeatother

\graphicspath{{figures/}}

\title{Developing a Series of AI Challenges for the United States Department of the Air Force}

\author[1,3]{Vijay Gadepally}
\author[1]{Gregory Angelides}
\author[2]{Andrei Barbu}
\author[4]{Andrew Bowne}
\author[1] {Laura J. Brattain}
\author[6] {Tamara Broderick}
\author[4]{Armando Cabrera}
\author[1]{Glenn Carl}
\author[4] {Ronisha Carter}
\author[1]{Miriam Cha}
\author[1] {Emilie Cowen}
\author[2]{Jesse Cummings}
\author[6]{Bill Freeman}
\author[2]{James Glass}
\author[7]{Sam Goldberg}
\author[6]{Mark Hamilton}
\author[5] {Thomas Heldt}
\author[6]{Kuan Wei Huang}
\author[6]{Phillip Isola}
\author[2]{Boris Katz}
\author[5]{Jamie Koerner}
\author[6]{Yen-Chen Lin}
\author[2]{David Mayo}
\author[4]{Kyle McAlpin}
\author[7]{Taylor Perron}
\author[1]{Jean Piou}
\author[1] {Hrishikesh M. Rao}
\author[1] {Hayley Reynolds}
\author[2]{Kaira Samuel}
\author[1]{Siddharth Samsi}
\author[7]{Morgan Schmidt}
\author[1]{Leslie Shing}
\author[1] {Olga Simek}
\author[4]{Brandon Swenson}
\author[6] {Vivienne Sze}
\author[1]{Jonathan Taylor}
\author[2]{Paul Tylkin}
\author[1] {Mark Veillette}
\author[1]{Matthew L Weiss}
\author[1]{Allan Wollaber}
\author[1] {Sophia Yuditskaya}
\author[1,3]{Jeremy Kepner}

\affil[1]{MIT Lincoln Laboratory}
\affil[2]{MIT Computer Science and Artificial Intelligence Laboratory}
\affil[3]{MIT Connection Science}
\affil[4]{United States Department of the Air Force}
\affil[5]{MIT Institute for Medical Engineering \& Science}
\affil[6]{MIT Electrical Engineering and Computer Science}
\affil[7]{MIT Department of Earth, Atmospheric and Plenetary Science}

\begin{document}

\maketitle

\begin{abstract}
Through a series of federal initiatives and orders, the U.S. Government has been making a concerted effort to ensure American leadership in AI. These broad strategy documents have influenced organizations such as the United States Department of the Air Force (DAF). The DAF-MIT AI Accelerator is an initiative between the DAF and MIT to bridge the gap between AI researchers and DAF mission requirements. Several projects supported by the DAF-MIT AI Accelerator are developing public challenge problems that address numerous Federal AI research priorities. These challenges target priorities by making large, AI-ready datasets publicly available, incentivizing open-source solutions, and creating a demand signal for dual use technologies that can stimulate further research. In this article, we describe these public challenges being developed and how their application contributes to scientific advances.
\end{abstract}

% sections
\section{Introduction}

% Potential organization:
% \begin{enumerate}
%     \item Climate Change (SEVIR, MultiEarth)
%     \item Health and Safety Systems (CogPilot, ManeuverID)
%     \item Cyber Security (Datacenter, RFChallenge (maybe), MagNav)
%     \item RL Environments or Decision Support? (AF Arcade, Race to Find Food (maybe)) 
%     \item Responsible AI (Puckboard, Spoken ObjectNet)
% \end{enumerate}

\begin{figure}[!htpb]
\centering
  \includegraphics[width=\textwidth]{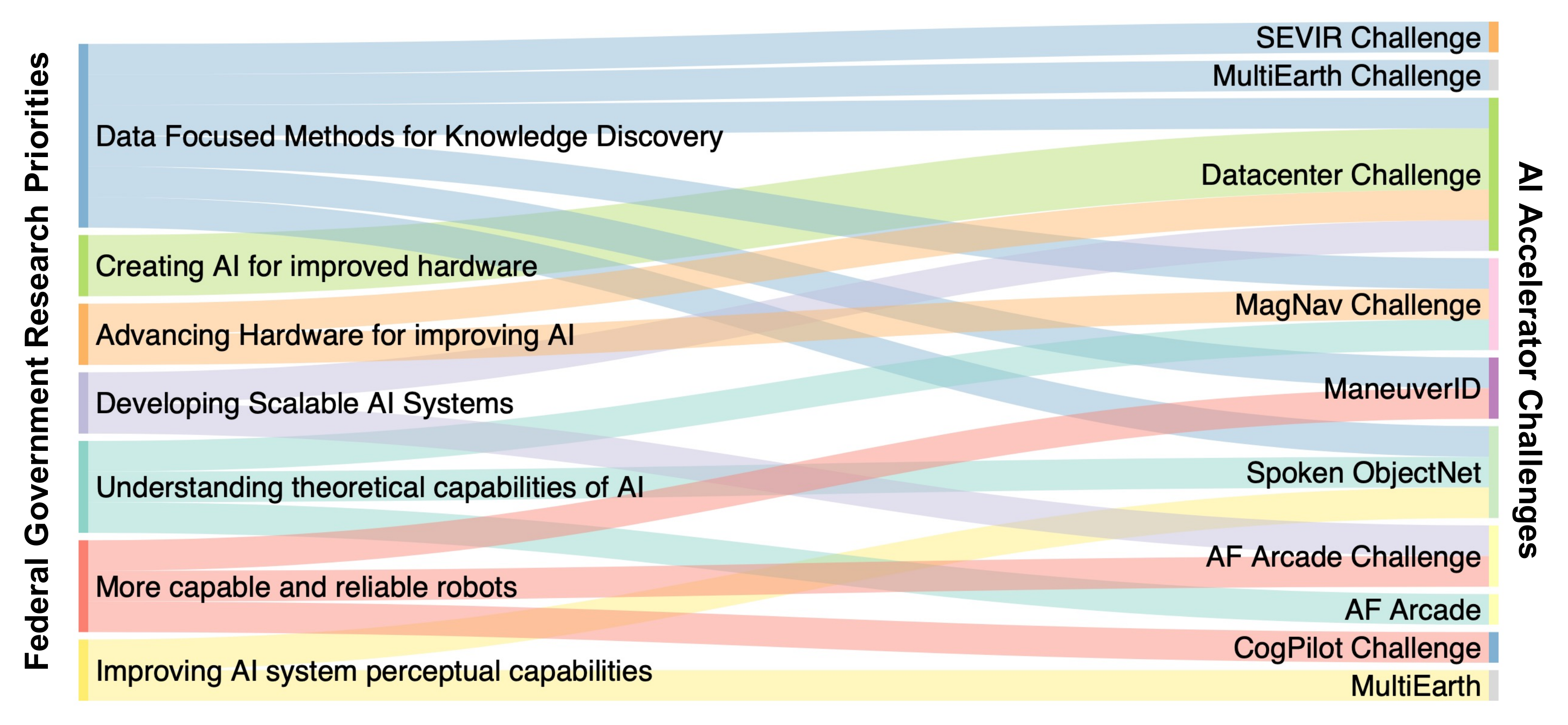}
  \caption{Mapping between selected Federal AI research priorities and DAF-MIT AI Accelerator Challenges. Federal research priorities are drawn from a number of AI strategy reports including ~\cite{national2019national}.}

  \label{fig:intro}
\end{figure}

In 2019, the United States Federal government outlined a broad strategy to ensure leadership in artificial intelligence through \href{https://www.ai.gov/}{Executive Order 13859}. Core to this strategy, outlined in ~\cite{national2019national}, are a number of research investments needed to achieve American AI leadership. Examples of these research thrusts are shown on the left side of Figure~\ref{fig:intro} (drawn from ~\cite{national2019national}).  Federal government organizations, such as the \href{https://media.defense.gov/2019/Feb/12/2002088963/-1/-1/1/SUMMARY-OF-DOD-AI-STRATEGY.PDF}{Department of Defense} and  \href{https://www.af.mil/Portals/1/documents/5/USAF-AI-Annex-to-DoD-AI-Strategy.pdf}{Department of the Air Force}, developed their own AI strategies to realize the broad goals of Executive Order 13859. One  initiative of the Department of the Air Force was the development of an Artificial Intelligence Accelerator hosted at MIT (\href{https://aia.mit.edu/}{DAF-MIT AI Accelerator}). The Accelerator consists of multidisciplinary teams from MIT and the Department of the Air Force to develop fundamental technologies that can advance AI within the Air and Space Forces and society in general. In order to drive innovation in relatively new fields and to fully engage with the wider AI community, the Accelerator is developing a series of ``challenge'' problems. These ``challenges'' consist of open datasets, benchmarks, problem definitions, metrics,  baseline implementations and address scientific problems such as weather prediction, datacenter optimization and human-machine interfaces. Developing multiple challenges in parallel has also supported synergistic activities. For example, we were able to develop reproducible pipelines for releasing open-source datasets, provide uniform computing platforms and leverage data created for one challenge for other challenges.

In this article, we describe a number of the challenges being developed through the DAF-MIT AI Accelerator with a focus on datasets and codebases that are currently released. Further, developing over 10 challenge problems has led to a number of important lessons which we describe. Beyond the challenges described in this article, the team is also developing additional challenges that will be released soon. For example, the \emph{Autonomous Flight Arcade} \citep{afarcade} is a suite of environments playable by both humans and AIs, that are designed for training artificial agents to accomplish complex tasks that are inspired by real-world aviation scenarios. 

\vspace{-5pt}
\subsection{Responsible Research Statement}
The various challenges and associated datasets included in this article come from a diverse set of problems that have diverse ethical considerations. While most projects do not have negative societal impact, every project undergoes a thorough security policy review. For projects that require data from human subjects, they undergo reviews from the Institutional Review Board at MIT (Committee on the Use of Humans as Experimental Subjects - COUHES) and if applicable, the Department of the Air Force (Human Research Protection Office - HRPO). Two projects outlined in this article underwent such reviews - the CogPilot Challenge (MIT/COUHES Protocol 2004000146, AFRL/HRPO Protocol: FWR20200141O) and the Spoken ObjectNet Challenge (MIT/COUHES Protocol 1508154186). The Department of Defense process for clearing human subject research is provided by 32 CFR 219, DoDI 3216.02 and the Office of Human Research Protections (OHRP) within the Department of Health and Human Services (DHHS). In cases in which consent was required, each dataset that required consent obtained it under both MIT's IRB (COUHES) and DAF's IRB (HRPO) approval process which requires ``Informed Consent.'' Only one of the projects (Spoken ObjectNet) leveraged crowdsourced data. The details of this process are outlined in Section 3 of ~\citep{spokenobjectnet}.

Additionally, we realize there are limitations to each of the works, and each challenge has worked to reduce the potential for bias in the data collected. A detailed description of these limitations is out of scope of this article since they are unique to each project. However, these limitations are highlighted in the references describing individual challenges. To the best of our knowledge, none of the datasets contain offensive content or personally identifiable information. In cases in which data sensitivity was possible, relevant fields were anonymized or de-identified. 

\section{Storm Event Imagery (SEVIR) Dataset}
\label{sec:sevir}

Adequately monitoring and forecasting Earth's weather is critical for environmental intelligence and human safety, especially in this unprecedented time of climate change.  Recently, AI has shown potential to improve the performance of forecasts by leveraging massive amounts of Earth Systems Datasets used in forecasting~\cite{schultz2021can}.  Geostationary satellite data provided by platforms like GOES, or rapidly updating radar information provided by NEXRAD, are essential for understanding current weather conditions, and for seeding forecasts.  However the size and complexity of these datasets can often be a hindrance to AI research, and unlike other AI sub-fields like computer vision and natural language processing, there are not many common benchmark datasets for the community to use for validating and benchmarking new capabilities.  

The Storm EVent ImageRy (SEVIR) dataset was created to address this issue \cite{veillette2020sevir}.  SEVIR combines a number of different weather sensing modalities, including geostationary satellite imagery, lightning detection and radar, into a single AI-ready dataset designed for studying several problems in meteorology.  SEVIR contains over 10,0000 events that each consist of 384 km x 384 km image sequences spanning 4 hours of time.  Many of the events in SEVIR were selected based on the \href{https://www.ncdc.noaa.gov/stormevents/}{National Center for Environmental Information Storm Event Database}.  
\begin{figure}[!htpb]
\centering
  \includegraphics[width=\linewidth]{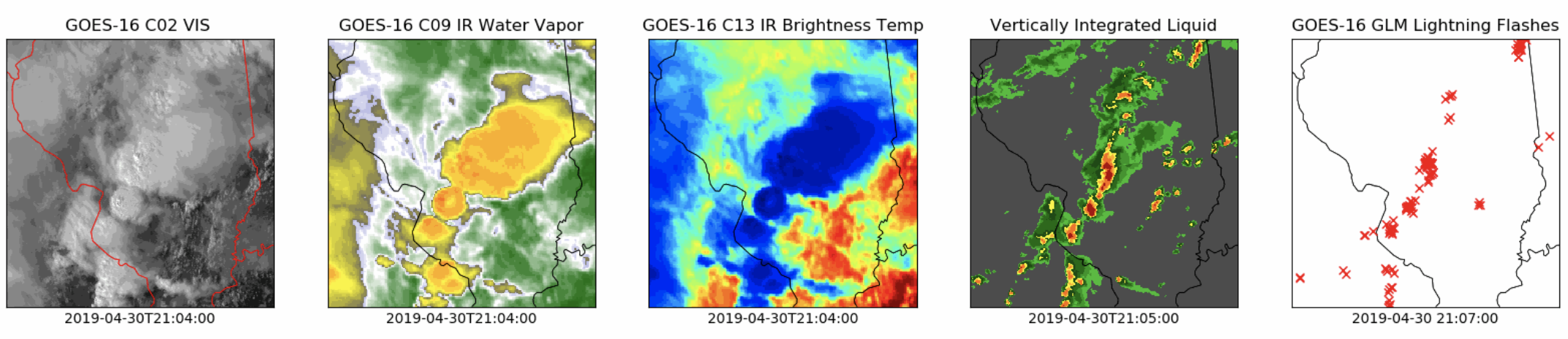}
  \caption{The  Storm EVent ImagRy (SEVIR) dataset contains over 10,000 spatially and temporally aligned sequences across the five image types. }
  \label{fig:sevir}
\end{figure}

\paragraph{Dataset \& Availability:}

SEVIR is a collection of temporally and spatially aligned image sequences depicting weather events captured over the contiguous US (CONUS) by the GOES-16 satellite and the mosaic of NEXRAD radars.  Figure~\ref{fig:sevir} shows a set of frames taken from a SEVIR event.  Each event in SEVIR consists of a 4-hour length sequence of images sampled in 5 minute steps.  The lightning modality is the only non-image type, and is represented by a collection of lightning flashes captured in the 4 hour time window by the GOES Geostationary Lightning Mapper (GLM).  SEVIR events cover 384 km x 384 km patches sampled at locations throughout CONUS.  The pixel resolution in the images differ by image type, and were chosen to closely match the resolution of the original data.  As of this writing, the SEVIR dataset can be obtained from \url{https://registry.opendata.aws/sevir/}. 
\paragraph{Challenges:}

The SEVIR dataset was constructed as a benchmark dataset for a number of problems in meteorology. The \emph{Nowcasting} task aims to develop short-term forecasts.  Due to the immense computation requirements of numerical weather prediction,  short term forecasts in the 0-2 hour range, commonly called nowcasts, are used for applications that require rapidly updating and high resolution forecasts.  A number of studies have proven deep learning to be effective at generating nowcasts~\cite{ravuri2021skilful}, and SEVIR provides a robust dataset for training and validating these models across a number of high impact weather events. The second task, \emph{Synthetic Weather Radar Generation}, is motivated by the desire to generate  "proxies" of radar-fields. Weather radar provides timely and detailed depictions of storm coverage, however radar is only available in small portions of the planet.   Synthetic weather radar models~\cite{veillette2018creating},\cite{hilburn2021development} generate synthetic "proxies" of radar-fields that can be used in areas lacking access to radar (e.g. over ocean).   SEVIR has aligned radar and non-radar modalities that can be used to train and validate these image-to-image models. The final task, \emph{Statistical Downscaling}, aims to address computational challenges of weather and climate models. Due to their large computational budget, current weather and climate models are run at spatial resolutions that are are too coarse to meet operational needs.  Radar and satellite data in SEVIR can be intentionally degraded to simulate low-resolution model output that is averaged over large spatial regions.  ML models can then be trained to restore features in the storms that were lost in this degradation.

\section{Datacenter Challenge}
\label{sec:dcc}
As AI/ML workloads become an increasingly larger share of the compute workloads in High-Performance Computing (HPC) centers  and  commercial  cloud  systems, there is a need to better understand cluster/datacenter operations. The Datacenter Challenge~\cite{dcc-dataset} aims to foster innovation in AI approaches to the analysis of large scale datacenter monitoring logs to reduce energy consumption, improve scheduling policies, optimize resource use, and identify policy violations.

\begin{figure}[!htpb]
    \centering
    \subfloat[GPU power draw]{\includegraphics[width=.3\linewidth]{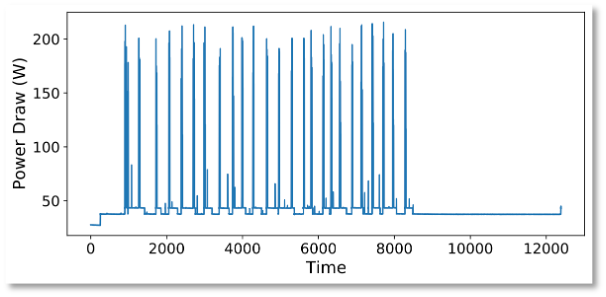}}
    \subfloat[GPU/CPU utilization]{\includegraphics[width=.4\linewidth]{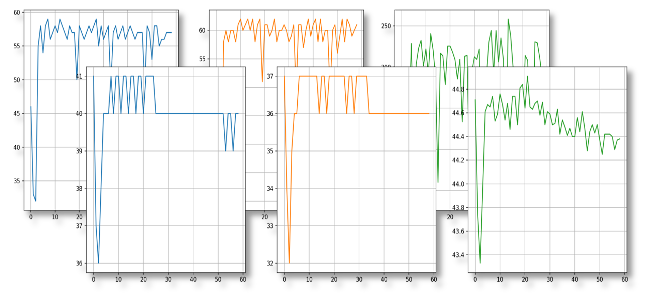}}
    \subfloat[GPU temperature]{\includegraphics[width=.3\linewidth]{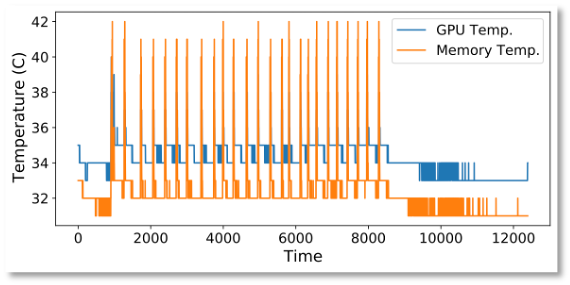}}
    \caption{Example time series data available as part of the MIT Supercloud dataset.}
    \label{fig:datacenter}
\end{figure}

% \begin{figure}[h]
% \centering
%     \subfloat[\centering GPU and CPU jobs runtime distribution and queue  wait  time  of jobs]{{\includegraphics[width=.45\linewidth]{figures/dcc-runtime.pdf}}}%
%     \qquad
%     \subfloat[\centering Distribution and total  GPU  hours  consumed  by different types of jobs]{{ \includegraphics[width=.42\linewidth]{figures/dcc-piechart.pdf} }}%
%     \caption{Initial analysis of the MIT Supercloud dataset~\cite{hpca22}}%
%     \label{fig:datacenter}%
% \end{figure}
    
% \begin{figure}[h]
%   \centering
%   \includegraphics[width=.45\linewidth]{figures/dcc-runtime.pdf}
%   \includegraphics[width=.43\linewidth]{figures/dcc-piechart.pdf}
%   \label{fig:datacenter}
% \end{figure}

\paragraph{Dataset \& Availability:}
The MIT Supercloud Dataset~\cite{dcc-dataset} was collected on the TX-Gaia system, which is a heterogenous cluster consisting of 224 nodes with two 20-core Intel Xeon Gold 6248 processors and two NVIDIA Volta V100 GPUs with 32GB of RAM each, and 480 nodes with Intel Xeon Platinum 8260 processors. The dataset consists of time series of CPU and GPU utilization, memory utilization, GPU temperature, snapshots of compute node state, file I/O, and the scheduler log. Examples of time series data included in the dataset are shown in Figure~\ref{fig:datacenter}. Currently over 2.1 TB of data are available for download at \url{https://dcc.mit.edu}.  Details of data collection, parsing and anonymization are available in ~\cite{dcc-dataset} and an initial analysis of the dataset in ~\cite{hpca22}.
%An initial analysis of this dataset focused on understanding system usage, with a focus on the usage patterns of GPUs~\cite{hpca22}.

\paragraph{Challenges:}
The Datacenter Challenge consists of two broad challenges. The first \emph{Workload Classification} challenge aims to characterize computing workloads. Understanding workload characteristics in HPC and Cloud systems is an important aspect of operating large datacenters. The  \emph{Workload Classification}  problem  focuses on  analyzing  a  large  time  series  dataset  of compute characteristics to develop classifiers for compute workloads~\cite{tang2022mit}.  Approaches to workload classification can be supervised or unsupervised and leverage a set of known workloads that can be used for training.  Given the large size of the dataset, innovations in applying machine learning to terabyte-sized time series data is also be presented as a related challenge question. The second, \emph{Carbon Reduction in AI} challenge, addresses the need for environmentally concious computing solutions. Given the ever-increasing energy demands for AI, the \emph{Carbon Reduction in AI} challenge will focus on reducing the carbon footprint of training and inference of AI models.  Carbon reduction in AI can be achieved through innovations in  hardware  as  well  as  algorithms. This challenge aims to address the climate impact of AI with the following over-arching goals: (1) Development of power-efficient approaches to AI training and inference to improve Petaflops/Watt performance, (2) data-efficient computing to reduce training resources, (3) informed Machine Learning to simplify ML using prior knowledge, (3) energy-efficient neural network design and (4) adaptive, energy-efficient data center management.

\section{MagNav challenge}

\label{sec:magnav}

Harnessing the magnetic field of the Earth for navigation has shown promise as a viable alternative to other navigation systems. Commercial and government organizations have surveyed the Earth to varying degrees of precision by collecting and storing magnetic field data as magnetic anomaly maps. The variation within these anomaly maps enables navigation in conjunction with a traditional inertial navigation system. This technique does not rely on any external communications, is available globally at all times and in all weather \cite{magnavbasis} and is also very difficult to jam.

\paragraph{Flight Data:}
For a detailed overview of the dataset collected to support this challenge please refer to the associated \href{https://github.com/MIT-AI-Accelerator/MagNav.jl/blob/master/readmes/datasheet_sgl_2020_train.pdf}{data-sheet}. The dataset contains approximately 40 hours of flight over Ottawa Canada and the surrounding regions on a geological survey aircraft. This Cessna equipped with multiple sensors in a variety of location enables the collection of clean data from the magnetometer mounted on the tail stinger as well as noisy data from within the cabin. The dataset has been made publicly available on Zenodo \cite{magnavdata} with additional metadata available at \url{https://github.com/MIT-AI-Accelerator/MagNav.jl}.

\paragraph{Challenges:}
One of the primary challenges with magnetic navigation is that the single magnetic field measurement from the magnetometer is comprised of several components: the magnetic field from the Earth's core field, anomaly field, and aircraft field. The total magnetic field measured at the magnetometer is a linear superposition of the magnetic fields of the vehicle and the Earth (with small contributions from sources arising from diurnal variation and space weather), and the magnetometer reports the scalar magnitude of the net magnetic field vector \cite{Tolles1955}. Given the scalar nature of the measurements it can be difficult to separate the magnetic anomaly field from the total measurement which is a required step for navigation. Due to this difficulty, the purpose of this challenge problem is to compute a compensation to remove the aircraft magnetic field from the sensor reading in order to perform magnetic navigation. Testing on the dataset shows that the Earth's magnetic field can be extracted from the total magnetic field using machine learning and that some techniques such as those in \cite{gnadtMagnavML} are highly effective at extracting a near-navigable magnetic field compensation. 

The \emph{Magnetic Compensation Challenge} dataset was developed to support a signal enhancement challenge problem that is further detailed at \url{https://magnav.mit.edu} and in \cite{magnavchallenge}. This primary challenge is to create a compensation that generates a mapping from a subset of in cabin uncompensated magnetometers to the compensated tail stinger magnetometer's measurements. A submission is judged in terms of Standard Deviation from the tail stinger for each held back flight segment. The secondary \emph{Navigation Challenge} is the creation of a full navigation solution which likely leverages an Extended Kalman Filter in order to convert the magnetic readings into a position in terms of LLA (Latitude, Longitude, Altitude) position as in \cite{gnadtMagnavML}. This task is judged in terms of mean square error as the average position error acts as the more important metric of interest once the problem has been converted into the position domain.

\section{MultiEarth Challenge}
The Multimodal Learning for Earth and Environment Challenge (MultiEarth 2022) is the first competition aimed at the monitoring and analysis of deforestation in the Amazon rainforest at any time and in any weather conditions. The goal of the Challenge is to provide a common benchmark for multimodal information processing and to bring together the earth and environmental science communities as well as multimodal representation learning communities to compare the relative merits of the various multimodal learning methods to deforestation estimation under well-defined and strictly comparable conditions. MultiEarth 2022 has three sub-challenges: 1) matrix completion, 2) deforestation estimation, and 3) image-to-image translation. Our challenge website is available at  \url{https://sites.google.com/view/rainforest-challenge}.

\paragraph{Dataset:}

Participants will receive a multimodal remote sensing dataset that consists of Sentinel-1, Sentinel-2, Landsat 5 and Landsat 8 as reported in Table~\ref{tab:data}. Sentinel-1 uses a synthetic aperture radar (SAR) instrument, which collects in two polarization bands: VV and VH. Sentinel-2, Landsat 5 and Landsat 8 use optical instruments, which measure spectral bands in the visible and infrared spectra. Detailed band designations for each sensor can be found in Google Earth Engine Data Catalog for \href{https://developers.google.com/earth-engine/datasets/catalog/COPERNICUS_S1_GRD}{Sentinel-1}, \href{https://developers.google.com/earth-engine/datasets/catalog/COPERNICUS_S2_SR}{Sentinel-2}, \href{https://developers.google.com/earth-engine/datasets/catalog/LANDSAT_LT05_C02_T1_L2}{Landsat 5}, and \href{https://developers.google.com/earth-engine/datasets/catalog/LANDSAT_LC08_C02_T1_L2}{Landsat 8}. In addition to the multimodal data, we release corresponding deforestation maps that are manually labeled using monthly mosaic satellite images from Planet \cite{planet}.

\begin{table*}[!htpb]
\footnotesize
\centering 
\begin{tabular}{c|c|c|c|c|c}
\hline
\textbf{Sensor} & \textbf{Time} & \textbf{Bands}                                                                                      & \textbf{Resolution (m) } & \textbf{\# Images} & \textbf{Link}\\ \hline \hline
Sentinel-1                     & 2014-2021           & VV, VH                                                                                              & 10                      & 859,637     & \begin{tabular}[c]{@{}c@{}} \href{https://rainforestchallenge.blob.core.windows.net/dataset/sent1_vh_train.zip} {\color{blue}{link1}} \\ \href{https://rainforestchallenge.blob.core.windows.net/dataset/sent1_vv_train.zip}{\color{blue}{link2}}   \end{tabular}   \\ \hline
Sentinel-2                     & 2018-2021           & \begin{tabular}[c]{@{}c@{}}B1, B2, B3, B4, B5,\\ B6, B7, B8, B8A, B9,\\ B11, B12, QA60\end{tabular} & 10                     & 5,395,569    & \begin{tabular}[c]{@{}c@{}} \href{https://rainforestchallenge.blob.core.windows.net/dataset/sent2_qa_b1-b4_train.zip}{\color{blue}{link1}} \\ \href{https://rainforestchallenge.blob.core.windows.net/dataset/sent2_b5-b12_train.zip}{\color{blue}{link2}}   \end{tabular}   \\ \hline
Landsat 5                       & 1984-2012           & \begin{tabular}[c]{@{}c@{}}SR\_B1, SR\_B2,  SR\_B3, \\ SR\_B4, SR\_B5, ST\_B6, \\ SR\_B7, QA\_PIXEL\end{tabular}                    & 30                     & 3,550,378     & \begin{tabular}[c]{@{}c@{}} \href{https://rainforestchallenge.blob.core.windows.net/dataset/landsat5_qa_b1-b3_train.zip}{\color{blue}{link1}} \\ \href{https://rainforestchallenge.blob.core.windows.net/dataset/landsat5_b4-b7_train.zip}{\color{blue}{link2}} \end{tabular}  \\ \hline
Landsat 8                       & 2013-2021           & \begin{tabular}[c]{@{}c@{}}SR\_B1, SR\_B2, SR\_B3, \\ SR\_B4, SR\_B5, SR\_B6, \\ SR\_B7, ST\_B10, QA\_PIXEL\end{tabular}           & 30                     & 2,172,574     & \begin{tabular}[c]{@{}c@{}} \href{https://rainforestchallenge.blob.core.windows.net/dataset/landsat8_qa_b1-b5_train.zip}{\color{blue}{link1}} \\ \href{https://rainforestchallenge.blob.core.windows.net/dataset/landsat8_b6-b10_train.zip}{\color{blue}{link2}}  \end{tabular} \\ \hline
\end{tabular}
\caption{Overview of the multimodal remote sensing dataset}
\label{tab:data}
\end{table*}

\paragraph{Challenges:}

The MultiEarth challenge consists of three sub-challenges. The Matrix Completion Sub-Challenge focuses on filling in spatial, temporal, and modality gaps in remote sensing data, especially gaps created by unfavorable lighting, weather conditions, or other atmospheric factors. The Deforestation Estimation Sub-Challenge looks to perform a binary classification to predict whether a region is deforested or not. Finally, the Image-to-Image Translation  Sub-Challenge aims to model a distribution of possible electro-optical (EO) image outputs conditioned on a SAR input image.

\section{Spoken ObjectNet Challenge}
\label{sec:objnet}
%purpose 

%An ability to describe or search visually-oriented information via natural language offers many useful applications for retrieving information from multimodal documents.  
In recent years there has been tremendous interest and research activity in %the area of 
multimodal information processing for problems such as image captioning and video retrieval.  A variety of datasets exist to support research in these areas, which typically consist of a repository of images or video clips %that are 
paired with either text or spoken captions. Unfortunately, many of these datasets contain intrinsic biases that the models trained on those datasets then learn, which in turn degrades their performance on real-world data.  For example, %most images and videos uploaded to websites are nicely lit, well-framed, and contain objects in their usual settings.  In turn, 
image captioning models are biased towards describing people on beaches as happy and image classification models don't recognize wolves outside of a snowy backdrop~\citep{objectrecognition}.
To address such issues, a large-scale crowd-sourced, bias-controlled object classification dataset called ObjectNet was created, consisting of a corpus of entirely new images instead of relying on those already uploaded to the Internet in some form \citep{objectnet}.  In turn, the Spoken ObjectNet (SON) corpus extends the ObjectNet corpus by collecting spoken descriptions of the ObjectNet images, and creating a series of challenge tasks for the corpus, as described below.

%Workers were asked to position a variety of household objects in a certain way against a specified background.  
%The viewpoint of the camera was also controlled.  
%In this way, ObjectNet has systematic controls in place for some of the biases that most other datasets exhibit.

\paragraph{Dataset \& Availability:}
Our spoken caption collections were crowd-sourced via Amazon Mechanical Turk (AMT), whereby workers were given an image and asked to record themselves as if they were describing the image to someone who could not see it.  Workers were told they could describe shapes, objects, locations, colors, and anything else of interest as they saw fit.  After each recording was completed, we ran several validation steps to ensure that the recording met our requirements~\citep{spokenobjectnet}. 
%Each recording had to pass three validation checks in order for the worker to proceed to the next image, based on duration, number of spoken words (via automatic speech recognition), and language model likelihood of the hypothesized word sequence~\cite{spokenobjectnet}.  
In total, we collected over 100,000 captions.  One spoken caption per image in ObjectNet was selected to form the Spoken ObjectNet-50k dataset, with a total of 50,273 samples.  48,273 are included in the training set, and 1,000 are included in both the validation and test sets. Additionally, to extend the dataset and enhance its challenge value, we selected a class-balanced subset of 20,159 images that will---when the data collection is complete---receive a full set of 5 captions for additional challenges such as image captioning. We call this subset Spoken ObjectNet-20k, or SON-20k.\footnote{Dataset is publicly available at \url{https://groups.csail.mit.edu/sls/downloads/placesaudio}, and code available at \url{https://github.com/iapalm/Spoken-ObjectNet}.}
Currently underway are efforts to produce accurate transcripts of the spoken captions.  These %transcripts 
will be %made 
available in future releases of the dataset, in order to enable text-based and % evaluations as well as 
speech-based evaluations.

\paragraph{Spoken ObjectNet Challenges:}
There are three challenges envisioned for the Spoken ObjectNet corpus.  The objective of the \emph{multimodal retrieval} task is to retrieve the appropriate image that corresponds to a given speech (or text) caption from a large unseen set of test image candidates.  The \emph{reverse retrieval} task involves retrieving the appropriate caption that corresponds to a given test image. The \emph{image captioning} challenge makes use of the SON-20k dataset containing five spoken captions/image.  The transcripts will be used to measure a BLUE score to evaluate the generated caption. The third challenge we plan to create will focus on a \emph{visual question answering (VQA)} task, leveraging the SON-20k subset.  We plan to crowd-source questions and answers by showing images and captions, and asking them to generate questions that are answerable from the image and the caption, and to highlight the location of the answer in the caption.  With such a question-answer collection, we can create a VQA challenge whereby a model must provide an answer to the questions about an image.  %We plan to also leverage the SON-20k subset for this challenge.
\looseness=-1
\section{Maneuver ID Challenge}
The Air Force continues to face a pilot shortage, in part because of the lack of infrastructure and methodology required to expedite the Undergraduate Pilot Training (UPT) process \cite{SAFCOPTNtechsummary}.  Pilot Training Next (PTN) is an experimental program responsible for pilot training education that is entering a new phase in undertaking this issue through the introduction of virtual reality (VR) flight simulators\cite{SAFCOPTNtechsummary}. Optimization of these simulators requires an increase in their training and testing capabilities across many fronts, including maneuver identification and scoring. An AI Challenge can be set up to gather solutions across the AI community and integrate them into the PTN curriculum. The Maneuver Identification Challenge was created to identify maneuvers from trajectory data to improve flight safety and pilot training.

\begin{figure}[!htbp]
%     \begin{minipage}{0.48\textwidth}
%      \centering
%      \includegraphics[width=\linewidth]{figures/TSVexample.jpg}
%      \caption{This is an example of the beginning rows of a TSV file from PTN containing parameter values for a single session in the flight simulator.}\label{fig:TSVexample}
%   \end{minipage}\hfill
   \begin{minipage}{0.45\textwidth}
     \centering
     \includegraphics[width=\linewidth]{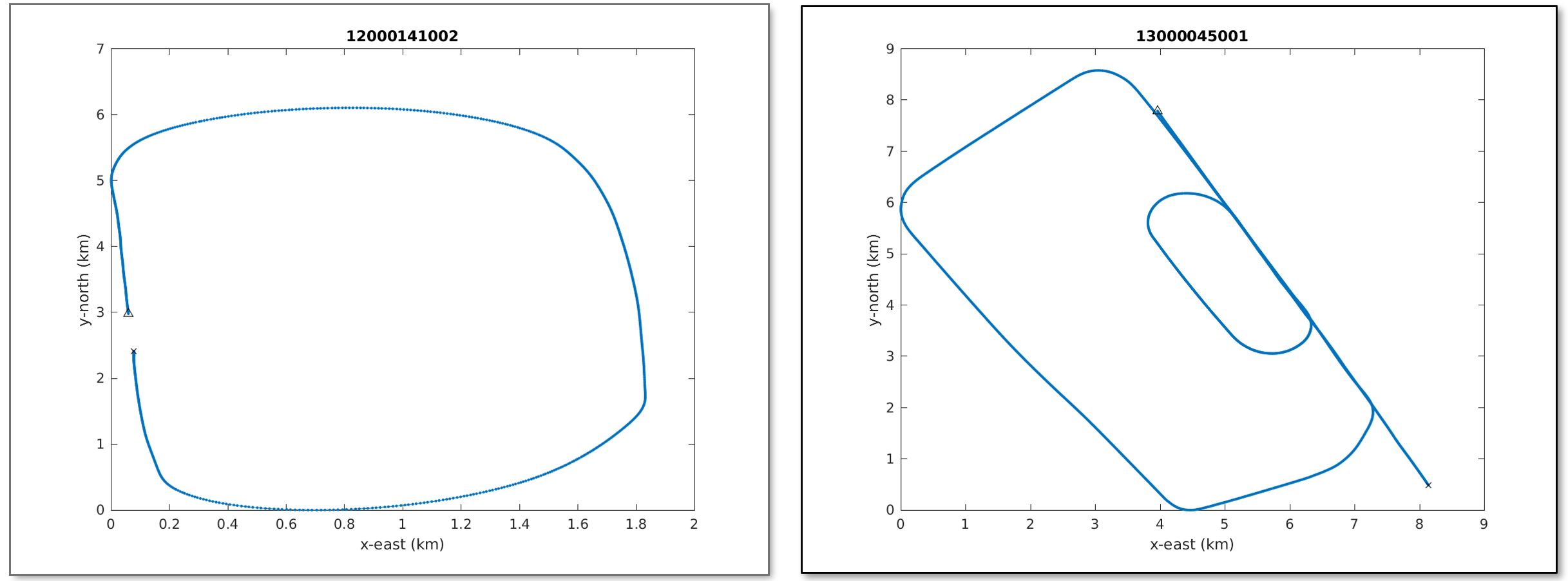}
     \caption{Two examples of PNG files showing an aerial view of the flight trajectory during the session. These are taken the first two simulator data sets provided by PTN.}\label{fig:PNGexample}
   \end{minipage}\hfill 
    \begin{minipage}{0.45\textwidth}
     \centering
     \includegraphics[width=0.9\linewidth]{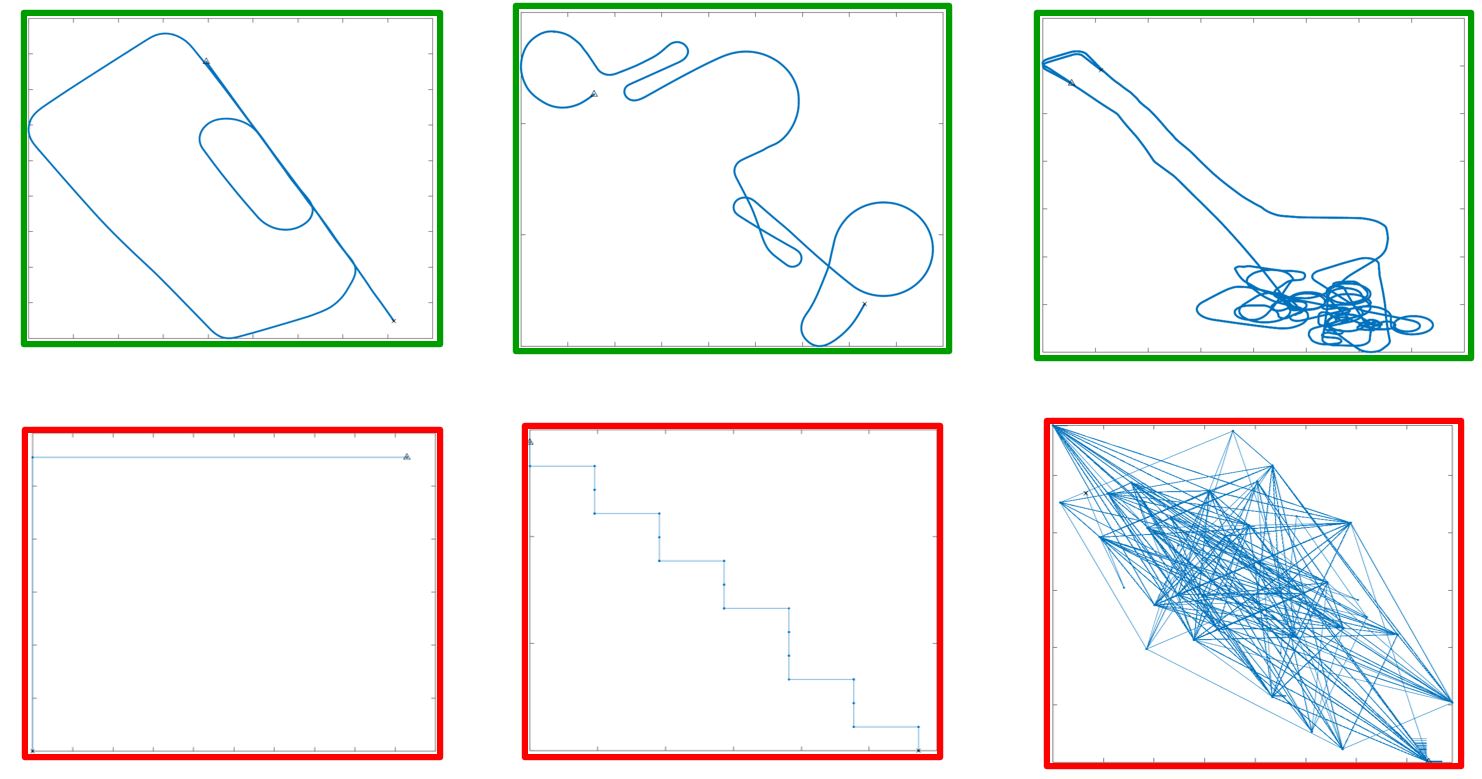}
     \caption{Figure shows examples of good data (green) and bad data (red), sorted manually. Bad data includes straight lines, jumps, and impossible maneuvers. }\label{fig:Challenge1}
   \end{minipage}  
\end{figure}

\paragraph{Data and Supplemental Resources:}
Available data from the VR simulators currently exist in both tab separated value (TSV) files and portable network graphics (PNG) files. TSV files consist of a plaintext file containing the time, position, velocity, and orientation of the flight and can be read by most data processing systems and viewed in any spreadsheet program. %Figure~\ref{fig:TSVexample} shows an example of a TSV file for a single session. 
PNG files contain images with an aerial view of the position of the aircraft during the session. Figure~\ref{fig:PNGexample} shows an example of a PNG file for a single session.

\paragraph{Challenges:}
There are three challenges that have been set up for researchers. The first deals with sorting the ‘good’ and ‘bad’ data into separate folders based on the presence of unbroken trajectories and identifiable maneuvers (good) and straight lines, jumps, and impossible maneuvers (bad). Examples of sorted data can be found in Figure~\ref{fig:Challenge1}.Currently, there is ‘truth data’ that has been sorted and verified manually, but is subject to change based on other factors that may affect the quality of the session. The second challenge deals with identifying which maneuvers the pilot  attempts to execute during their flight simulator run. No truth data for this challenge currently exists. The third challenge deals with scoring the pilot’s attempt once the maneuver has been identified. This could greatly benefit the efficiency and quality of the pilot training education. No truth data for this challenge currently exists.

% \begin{figure*}[htb]
%   	\centering
%     	\includegraphics[width=0.5\columnwidth]{figures/Challenge1.jpg}
% 	\caption{This figure shows examples of good data (green) and bad data (red), sorted manually. The bad data includes straight lines, jumps, and impossible maneuvers.}
%       	\label{fig:Challenge1}
% \end{figure*}

\section{CogPilot Challenge}
\label{sec:CPC}
Quantitative performance measurements and physiological monitoring are hypothesized to provide a more individualized and objective assessment of pilot training than current subjective, coarse measures. The CogPilot team seeks to develop AI algorithms to predict individual cognitive state and operational performance using multimodal physiological signals, thereby supporting personalized optimization of pilot training.

%\begin{figure}[h]
%    \centering
%    \subfloat[ Flight Simulator and Data Collection Setup]{\includegraphics[width=.45\linewidth]{figures/CogPilot_Fig1AB.png}}
%    \subfloat[ILS landing scenarios and mesures of flight performance]{\includegraphics[width=.45\linewidth]{figures/CogPilot_Fig2.png}}
 %   \caption{}
%    \label{fig:setup}
%\end{figure}

\begin{figure}[!htpb]
    \centering
    \includegraphics[width=\textwidth]{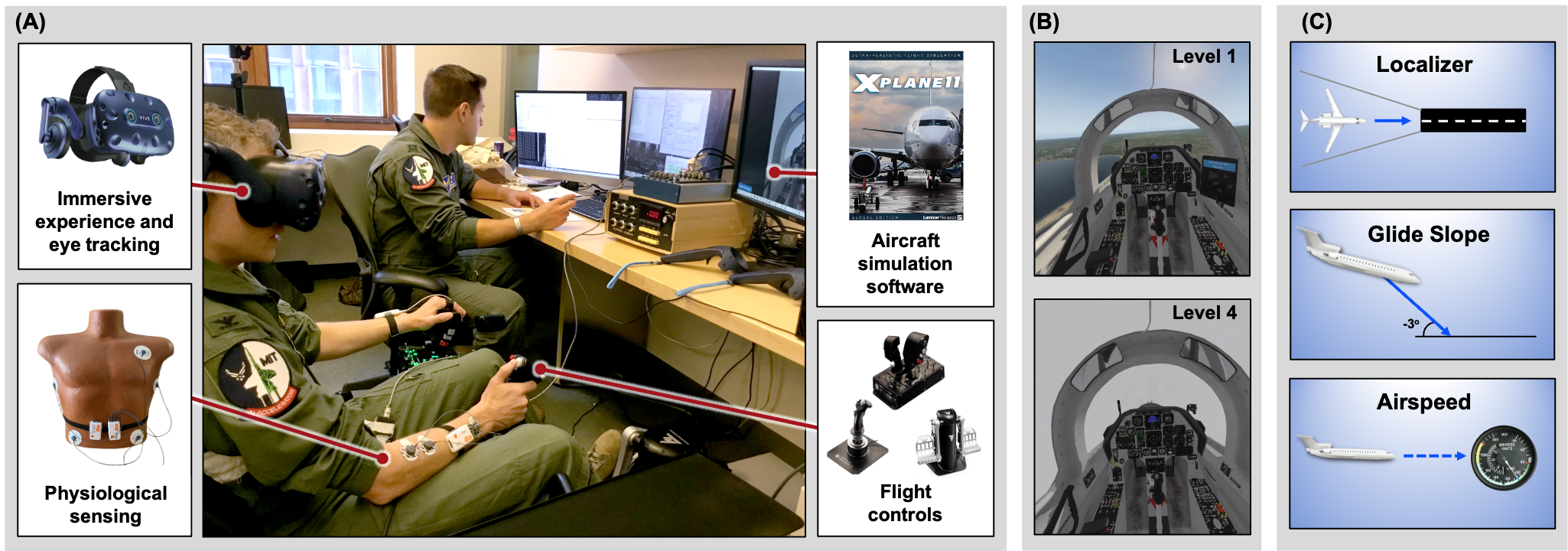}
    \caption{CogPilot challenge key components. (A) Flight simulator and data collection setup. (B) Example ILS landing scenarios. (C) Measures of flight performance.}
    \centering     
    \label{fig:setup} 
    \end{figure}

% \begin{figure}[h]
    
%     \includegraphics[width=0.9\textwidth]{figures/CogPilot_Fig2.png}
%     \caption{Example ILS landing scenarios and data challenge measures of flight performance}
%     \centering
%     \label{fig:performance}
% \end{figure}

\paragraph{Dataset Description and Availability:}
We have created an immersive virtual reality (VR) training environment simulating a real-life undergraduate Air Force pilot training facility (Figure~\ref{fig:setup}A). This setup allows us to replicate the pilot training experience while augmenting it to support multimodal data collection, including eye tracking, voice, and physiological measurements. The VR immersion is implemented using an HTC Vive Pro Eye headset with built in eye tracking capability. The simulation software engine is X-Plane 11, with the T-6A Texan II fixed wing trainer aircraft as the model aircraft. The simulated aircraft is flown using a pendular rudder and hands-on throttle-and-stick. The full simulation presents a series of Instrumented Landing Scenario (ILS) runs across four levels of difficulty (Figure~\ref{fig:setup}B) During each training run, the aircraft must be landed while maintaining tight lateral tolerances, elevation tolerances, and airspeed (velocity) tolerances. During a flying session, a subject performs 12 ILS runs with three iterations of each difficulty level. Flight performance is quantified as horizontal and vertical deviation from an ideal line towards the runway and deviation from the prescribed ideal speed (Figure~\ref{fig:setup}C). Physiological signals were aggregated and synchronized using Lab Streaming Layer. The complete dataset, together with data dictionary and python starter code, can be downloaded from \href{https://physionet.org/}{PhysioNet}.

\paragraph{Challenges:}
This data challenge seeks to explore how quantitative performance measurements and multimodal physiological data can provide individualized and more accurate assessment of a student pilot’s competency than current subjective, coarse measures. This additional level of cognitive state estimation can then be used to refine an individual’s training curriculum. The data was divided into a challenge set and an evaluation set by subject. Participants are encouraged to develop AI-based solutions for the two tasks by using physiological measurements (e.g., eye tracking, Electrocardiogram, Electromyography etc.) provided in the challenge dataset. \emph{Task} 1 is to classify which of the 4 difficulty levels the pilot was accomplishing using only physiological metrics. The grading metric is Area Under the ROC Curve classification performance between predicted difficulty level and true difficulty level. \emph{Task 2} is to estimate the pilot's overall deviation from desired lateral, vertical, and velocity targets using only physiological metrics. Task 2 is graded using Root Mean Squared Error between predicted deviations and true deviations.

% \textbf{Task 1:}  Predict difficulty level of a pilot training simulation run
 
% Metric: Compare the predicted level against the actual difficulty level using Area Under the Curve
 
% \textbf{Task 2:}  Estimate flight performance error
	 
% Metric: Compute Root Mean Squared Error between predictions and actual truth values

\section{Lessons Learned}

\begin{figure}[!htbp]
%\begin{wrapfigure}{l}{0.9\textwidth}
  	\centering
    	\includegraphics[width=\columnwidth]{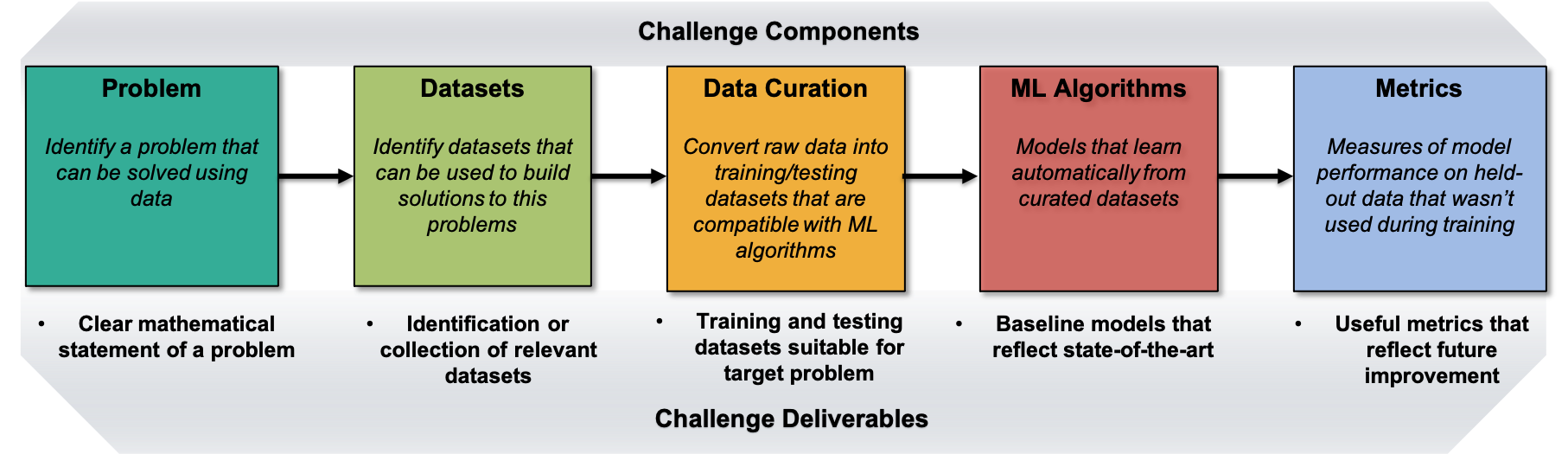}
	\caption{Components and description of challenges.}
      	\label{fig:components}
%\end{wrapfigure}
\end{figure}

Developing a challenge goes beyond just collecting and releasing data. For example, Figure~\ref{fig:components} describes other components required for an AI challenge. In our experience, developing a compelling problem, performing the expensive data curation steps~\cite{kindi2021technical,gadepally2015using}, and providing baseline implementations along with clear success metrics are very important steps to engaging the public.

The DAF-MIT AI Accelerator utilizes challenge problems to publicize data, engage the public, and advance the science of machine learning. Consistent with the National AI Research and Development Strategic Plan, the mission of the DAF-MIT AI Accelerator is establishing and building a sustained long-term research ecosystem to drive discovery and insight ~\cite{national2019national}. While the focus is on fundamental research, public challenges accelerate the transition of basic and advanced research to applied research directions that can lead to commercialization and operational capabilities. The DAF-MIT AI Accelerator focuses on those problems with dual-use -- problems that are important to the public, relevant to national defense -- and unlikely to be funded by investment-backed technology companies. Until novel machine learning capabilities are matured and proven out, the lack of commercial promise will leave critical needs unfulfilled by industry ~\cite{finalreport}. Applying research algorithms such as robustness and data augmentation to real problems fills this critical gap and accelerates the transition to commercialization and operational integration of AI. By some measures, ``82\% of the algorithms in use today originated from federally funded non-profits and universities” ~\cite{finalreport} - a measure which underscores the need for developing public challenge problems. 

% This pathway is proven; despite billions of dollars invested by private companies in machine learning, “82 percent of the algorithms in use today originated from federally funded non-profits and universities” ~\cite{finalreport}. Moreover, without injecting research into the ecosystem, the government will play a passive role in the development of AI ~\cite{finalreport}.

We have learned a number of valuable lessons from developing these AI challenges: 

\begin{itemize}
    \item \textbf{Development Platform:} Developing an AI challenge requires significant computational resources (especially for data curation and baseline algorithm development). As an example, each of the challenges outlined in this article leveraged the MIT SuperCloud~\cite{reuther2018interactive} for the upfront development. The development platforms are often distinct from the deployment and most projects deploy their challenge on public or private cloud platforms.
    \item \textbf{Data/Code Release:} Most organizations have an unwieldy review process for open-sourcing code and/or datasets. In our experience, including legal and contracting representatives from the onset is critical to streamline a continual release process for initial and updates to data and code. For example, each of the challenges outlined closely interacted with Air Force Judge Advocates (\url{https://www.afjag.af.mil/}) to identify any legal issues that may delay data/code releases. In cases where public release is not possible due to data sensitivity, for example in the ManeuverID Challenge, we developed template \href{http://www.mit.edu/~kepner/AI-Accelerator/DataSharingAgreement-Signable.docx}{Data Use Agreements} that data owners can tailor for their project. \looseness=-1 
    \item \textbf{Reproducible Pipelines:} In our experience, there are many small decisions that are made in the development of a challenge. For example, library dependencies, container technologies supported, data formats that have a large upstream consequences. It is important to handpick and document choices in order to improve reproducibility.
    \item \textbf{Engaging user community:} Each of the challenges outlined in this article has diverse scientific communities. It is important to bring this community into the challenge problem development as early as possible through workshops and technical exchanges. This ensures interesting problems, relevant metrics and baseline implementations that accurately reflect the field's state-of-the-art.

\end{itemize}

% The DAF-MIT AI Accelerator’s public challenges present open data, rigorously stated problems, and use cases that indicate return on investment. This approach sparks interest, and private funding, into real problems that have the potential to save or improve lives ~\cite{finalreport}. Challenging problems and opportunities to support national security are very attractive to academia and industry ~\cite{industry}. Thus, public challenges leverage the triad of AI research and development--government, academia, and industry--to full effect in advancing science to remain competitive in defense and economics.

\section{Summary}

The DAF-MIT AI Accelerator was created with the vision of bringing fundamental AI advances to the Department of the Air Force while also addressing societal needs. Part of this vision is realized through the development of public challenge problems in a variety of critical technical domains. This article describes a number of public challenges as well as important lessons learned from developing and deploying nearly a dozen AI challenges.

\newpage
\section*{Acknowledgements}
The authors wish to acknowledge the following individuals for their contributions and support: Bob Bond, Tucker Hamilton, Mike Kanaan, Shannon Mann, Sanjeev Mohindra, Charles Leiserson, Christian Prothmann, John Radovan and Daniela Rus,

Research was sponsored by the United States Air Force Research Laboratory and the United States Air Force Artificial Intelligence Accelerator and was accomplished under Cooperative Agreement Number FA8750-19-2-1000. The views and conclusions contained in this document are those of the authors and should not be interpreted as representing the official policies, either expressed or implied, of the United States Air Force or the U.S. Government. The U.S. Government is authorized to reproduce and distribute reprints for Government purposes notwithstanding any copyright notation herein.

\newpage
% \input{sections/checklist}
% \newpage
% \input{sections/appendix}
% \newpage

\bibliographystyle{unsrt}%{plainnat}
\bibliography{references}

\end{document}